\documentclass[a4paper]{cas-dc}

\usepackage[numbers]{natbib}

\def\tsc#1{\csdef{#1}{\textsc{\lowercase{#1}}\xspace}}
\tsc{WGM}
\tsc{QE}
\tsc{EP}
\tsc{PMS}
\tsc{BEC}
\tsc{DE}

\begin{document}
\def\floatpagepagefraction{1}
\def\textpagefraction{.001}
\shorttitle{Query-Conditioned Knowledge Alignment}
\shortauthors{Y. Jiao et al.}
\title [mode = title]{Query-Conditioned Knowledge Alignment for Reliable Cross-System Medical Reasoning}                      

\author[1]{Yan Jiao}[orcid=0009-0000-9028-4614]
\ead{yanyanjiao2018@gmail.com}

\author[1]{Jingran Xu}
\ead{17348800332@163.com}

\author[1,2]{Pin-Han Ho}[orcid=0000-0002-0717-1481]
\cormark[1]
\ead{pinhanho71@gmail.com}

\author[3,1]{Limei Peng}[orcid=0000-0001-9984-9861]
\ead{auroraplm@knu.ac.kr}

\affiliation[1]{organization={Shenzhen Institute for Advanced Study, 
University of Electronic Science and Technology of China},
                city={Shenzhen},
                country={China}}

\affiliation[2]{organization={Department of Electrical and Computer Engineering, 
University of Waterloo},
                city={Waterloo},
                country={Canada}}

\affiliation[3]{organization={School of Computer Science and Engineering, 
Kyungpook National University},
                city={Daegu},
                country={South Korea}}

\cortext[1]{Corresponding author}

\begin{abstract}
Cross-domain knowledge alignment is essential for integrating heterogeneous medical systems, yet existing approaches typically treat entity alignment as a static matching problem, ignoring query context and cross-system asymmetry.
This limitation is particularly critical in integrative medical settings, where correspondence between concepts is inherently context-dependent, non-bijective, and direction-sensitive.

In this paper, we propose Query-Conditioned Entity Alignment (QCEA), which reformulates entity alignment as a query-conditioned correspondence problem.
Instead of learning a fixed mapping between entity representations, QCEA treats the textual description of a source entity as a query and ranks candidate entities in the target graph, enabling context-dependent alignment.
The framework integrates semantic encoding, graph-based representation learning, and a direction-aware transformation module to capture asymmetric and many-to-many correspondence across heterogeneous knowledge systems.

We evaluate QCEA on TCM--WM knowledge graphs derived from SymMap, covering both symptom alignment and herb--molecule alignment tasks.
Experimental results show consistent improvements over representative baselines, particularly on rank-sensitive metrics such as Hit@K and MRR.
Furthermore, downstream retrieval-augmented generation (RAG) experiments demonstrate that improved alignment leads to better evidence retrieval, stronger grounding, and higher answer accuracy.
These findings highlight that alignment is not merely a data integration step, but a key factor that shapes knowledge accessibility and reliability in cross-system medical reasoning.
\end{abstract}

\begin{highlights}
\item Reformulates entity alignment as query-conditioned ranking
\item Handles asymmetric and many-to-many cross-system correspondence
\item Captures context-dependent alignment in medical knowledge graphs
\item Improves top-rank alignment under semantic ambiguity
\item Supports grounded retrieval and downstream medical reasoning
\end{highlights}

\begin{keywords}
Entity alignment \sep 
Knowledge graphs \sep 
Query-conditioned ranking \sep 
Cross-system medical knowledge integration \sep 
Retrieval-augmented generation
\end{keywords}
\maketitle
\section{Introduction}\label{sec:introduction}

Large language models (LLMs) have become increasingly important for medical reasoning and knowledge-intensive applications, where retrieval-augmented generation (RAG) improves factuality by grounding outputs in external knowledge sources~\cite{lewis2020retrieval}.
However, their effectiveness critically depends on the quality of retrieved evidence, which in practice requires consistent and well-aligned knowledge across heterogeneous medical systems.
Such heterogeneity, arising from distinct conceptual frameworks across clinical traditions, poses fundamental challenges for cross-system reasoning and retrieval~\cite{nicholson2020constructing,xue2024biomedical}.

An illustrative case is the integration of traditional Chinese medicine (TCM) and Western medicine (WM).
These paradigms adopt fundamentally different conceptual frameworks~\cite{matos2021understanding}, leading to semantic discrepancies that hinder cross-system interoperability and reduce the reliability of retrieval and clinical decision support~\cite{abu2023healthcare,sutton2020overview}.
With the growing adoption of integrative medicine~\cite{world2019global}, aligning heterogeneous medical knowledge systems has become increasingly important.

Beyond interoperability, such alignment is crucial for knowledge-driven reasoning systems.
In modern LLM-based pipelines, RAG relies on aligned knowledge to provide grounded evidence for downstream inference.
However, when cross-system correspondence is ambiguous or misaligned, errors in alignment propagate into incorrect evidence selection, resulting in ungrounded or hallucinated outputs.
Therefore, accurate cross-system alignment is not merely a data integration problem, but a prerequisite for reliable knowledge grounding and reasoning in heterogeneous medical artificial intelligence (AI) systems.

\begin{figure}
\centering
\includegraphics[width=\columnwidth]{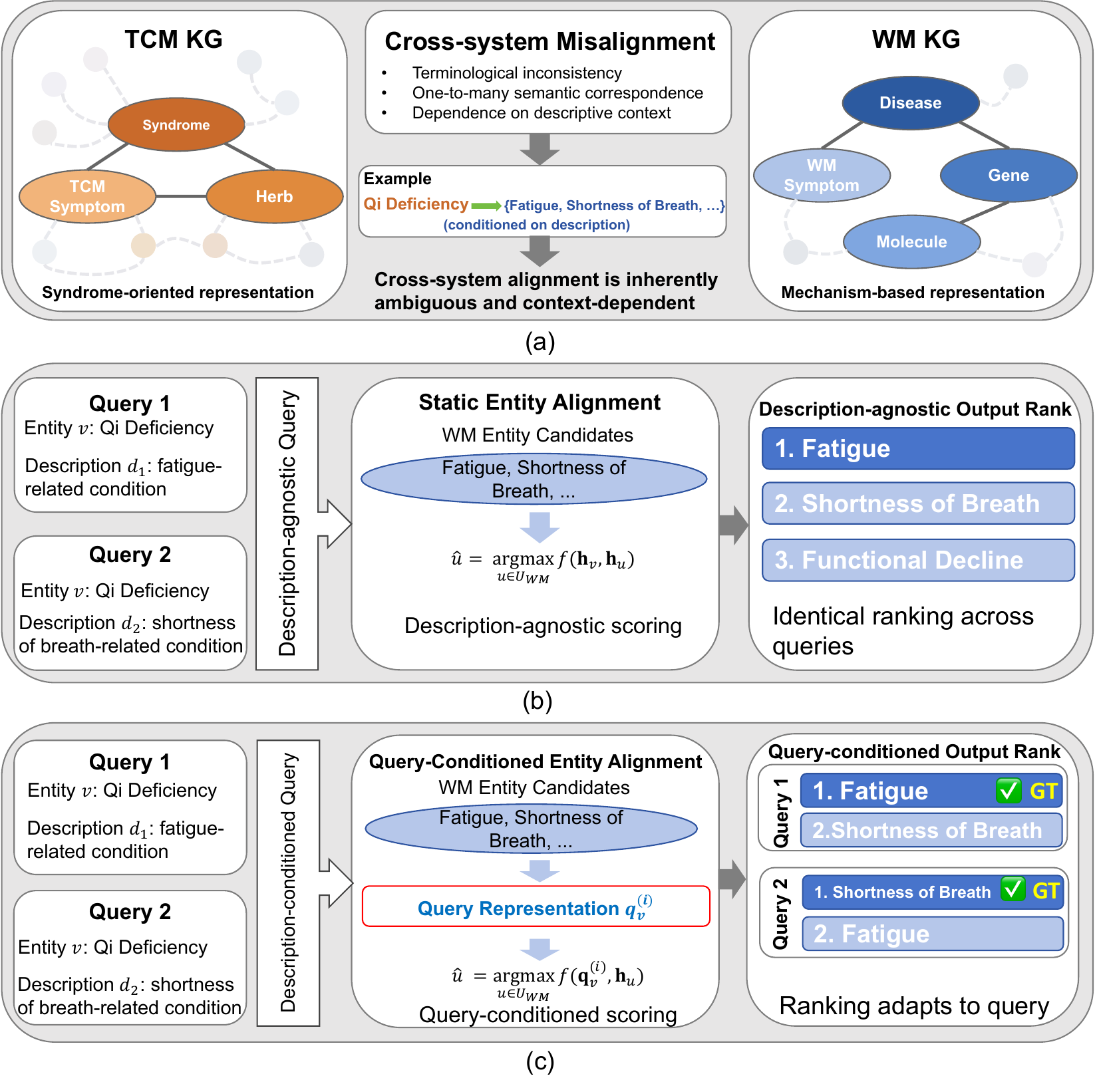}
\caption{Cross-system semantic misalignment and query-conditioned entity alignment.
(a) Context-dependent ambiguity in cross-system correspondence.
(b) Description-agnostic entity-based alignment produces identical rankings for different descriptions.
(c) QCEA enables query-conditioned alignment with context-dependent outputs.
}
\label{fig:framework}
\end{figure}

A central challenge is the ambiguous and context-dependent nature of entity correspondence.
In TCM, a single clinical concept can correspond to different WM entities depending on its descriptive context.
For example, \textit{Qi Deficiency} may align with fatigue-related or shortness-of-breath-related manifestations under different descriptions.
As illustrated in Fig.~\ref{fig:framework}(a), such correspondence is inherently context-dependent.
However, as shown in Fig.~\ref{fig:framework}(b), existing methods based on fixed entity representations and static similarity functions produce identical candidate rankings for different descriptions, failing to capture such variability.

Furthermore, cross-system correspondence is inherently non-bijective and direction-asymmetric.
For example, a TCM syndrome such as \textit{phlegm-heat obstruction in the lung} may correspond to multiple WM symptoms, while a WM symptom such as cough may relate to multiple TCM syndromes.
Such many-to-many and asymmetric relations are not well modeled by approaches that assume symmetric or one-to-one matching, highlighting the need for more flexible alignment mechanisms.

These limitations conflict with the requirements of LLM-based retrieval and reasoning, where relevance is inherently query-dependent.
However, most existing entity alignment approaches optimize fixed pairwise matching objectives, implicitly assuming that correspondence remains invariant across contexts.
This assumption creates a mismatch with downstream retrieval, in which the relevance of a candidate entity depends on how the query is expressed.
Existing methods, including structural approaches based on graph topology~\cite{hao2021medto}, embedding-based models with static similarity functions~\cite{zhang2019multike}, and text-enhanced methods~\cite{zhang2023autoalign}, perform alignment over fixed entity pairs, which limits their ability to capture context-dependent correspondence.

We propose \textbf{Query-Conditioned Entity Alignment (QCEA)}, which reformulates entity alignment as a query-conditioned ranking problem.
QCEA treats the textual description of a source entity as a query and ranks candidates in the target graph, aligning the alignment objective with query-dependent relevance in retrieval.
Unlike standard dense retrieval, where queries and candidates are predefined and the correspondence relation is typically many-to-one or relevance-based, our setting requires alignment under asymmetric, non-bijective, and direction-dependent cross-system correspondence.
The framework integrates domain-specific semantic encoding, graph-aware representation learning, a direction-aware Tucker projection, and a many-to-many contrastive objective to model context-dependent, asymmetric, and non-bijective correspondence.
As shown in Fig.~\ref{fig:framework}(c), different descriptions of the same entity can lead to different ranking outcomes, enabling context-sensitive alignment and improving cross-system retrieval quality for LLM-based reasoning.

We evaluate QCEA on heterogeneous medical knowledge graphs from the respiratory subgraph of SymMap~\cite{wu2019symmap}. This evaluation focuses on TCM symptom–WM symptom alignment and herb–molecule alignment tasks.

QCEA improves performance over representative baselines, particularly on rank-sensitive metrics such as Hit@K and mean reciprocal rank (MRR).
We also examine the downstream impact of alignment quality in LLM-based RAG settings, where improved alignment leads to better cross-system retrieval, stronger answer grounding, and higher question-answering accuracy.

Overall, QCEA serves as a unified framework that integrates semantic and structural signals for context-aware alignment, enabling more reliable cross-system retrieval and reasoning in heterogeneous medical knowledge graphs.

\paragraph*{Key novelty.}
The contribution of QCEA does not lie in introducing new model components, but in redefining entity alignment as a query-conditioned correspondence problem.
This shift changes the objective from learning a fixed mapping to modeling a conditional relation, which is essential in heterogeneous medical systems where the same source concept may correspond to different target concepts depending on descriptive context.
\vspace{4pt}

This work makes the following contributions:
\begin{itemize}
\item \textbf{Query-conditioned formulation of entity alignment:}
We reformulate cross-domain entity alignment as a retrieval-oriented problem that conditions correspondence on query semantics.
Rather than claiming novelty in query-conditioned ranking itself, we focus on adapting this paradigm to settings with asymmetric and many-to-many cross-system mappings.

\item \textbf{Direction-aware alignment modeling:}
We introduce a direction-aware transformation framework that captures the asymmetry between heterogeneous medical systems, enabling distinct alignment behaviors across domains.

\item \textbf{Comprehensive evaluation with downstream validation:}
We evaluate the proposed approach on TCM--WM datasets and demonstrate consistent improvements in alignment accuracy.
Additionally, downstream RAG experiments show that alignment quality consistently affects retrieval quality, answer accuracy, and evidence grounding.
\end{itemize}

The remainder of this paper is organized as follows. 
Section 2 reviews related work on knowledge graph alignment, semantic matching, and medical knowledge integration.
Section 3 presents the proposed QCEA framework, including query representation, graph encoding, direction-aware projection, and the many-to-many contrastive objective.
Section 4 describes the experimental setup, datasets, baselines, and evaluation protocols.
Section 5 reports experimental results, ablation studies, and downstream RAG evaluation.
Section 6 discusses implications and limitations, and Section 7 concludes the paper.

\section{Literature Review}

Cross-system medical knowledge integration involves aligning heterogeneous conceptual systems to support reliable retrieval and reasoning.
In settings such as TCM and WM, discrepancies in terminology and abstraction create challenges for consistent knowledge grounding~\cite{nicholson2020constructing,xue2024biomedical}.
Existing research spans entity alignment, semantic matching, and medical knowledge integration, but these directions remain loosely connected, particularly in handling context-dependent and asymmetric correspondence.

\subsection{Knowledge Graph Entity Alignment}

Knowledge graph entity alignment focuses on identifying corresponding entities across heterogeneous graphs by learning representations that preserve cross-graph similarity~\cite{zhu2024survey}.
Existing approaches differ mainly in how such representations are constructed.
Structural methods leverage graph topology to capture relational consistency~\cite{wang2018cross,wu2019relation,yang2019aligning}, embedding-based models learn unified latent spaces for alignment~\cite{sun2017cross,chen2017multilingual,zhang2019multike,sun2018bootstrapping,xiang2021ontoea}, and more recent efforts explore improving representation quality through contrastive learning, multi-view modeling, and the use of textual features~\cite{cheng2025easyea,yang2025daea}.

Despite these differences, most existing methods formulate alignment as a fixed pairwise similarity function between entity representations, i.e., a context-independent mapping between entities once embeddings are learned.
This assumption is often violated in cross-system medical settings, where correspondence depends on how entity semantics are described and may vary across contexts, leading to asymmetric and many-to-many relationships.
As a result, fixed pairwise matching is insufficient to capture the variability of cross-system correspondence.
In addition, these approaches are not naturally aligned with retrieval pipelines, where candidate relevance is conditioned on a query and directly affects downstream reasoning.

\subsection{Query-Conditioned Semantic Matching and Retrieval}

Semantic matching in information retrieval is commonly framed as a query-conditioned ranking problem, with representative approaches including DRMM, DPR, and ColBERT~\cite{guo2016deep,karpukhin2020dense,khattab2020colbert}.
These methods are widely adopted in RAG systems, where they are used to retrieve evidence conditioned on the input query for downstream generation.

However, these models operate within a single semantic space and do not explicitly model correspondence across heterogeneous conceptual systems.
While they capture query-dependent relevance, alignment additionally requires resolving cross-system ambiguity, where mappings may be asymmetric or one-to-many. 
Consequently, even when conditioned on queries, retrieval models alone cannot ensure consistent access to aligned knowledge across systems.

While RAG~\cite{lewis2020retrieval} improves downstream reasoning, its effectiveness depends on reliable access to relevant knowledge across sources.
In cross-system settings, this requirement places alignment as a prerequisite for effective retrieval.
This motivates approaches that extend query-conditioned ranking with mechanisms for modeling cross-system correspondence.

\subsection{Medical Knowledge Integration and LLM-based Applications}

Heterogeneous medical knowledge integration focuses on harmonizing multi-source biomedical data to support applications such as clinical decision support and knowledge retrieval~\cite{xu2025survey}.
In TCM--WM settings, discrepancies extend beyond terminology to fundamental differences in conceptual systems and diagnostic abstractions~\cite{sun2013differences,yu2023developing}.

Resources such as SymMap~\cite{wu2019symmap} provide cross-system links between TCM and biomedical entities, yet these mappings are frequently incomplete and context-dependent, and are typically constructed offline without adapting to query-specific semantics. 
Terminology-based frameworks such as UMLS~\cite{bodenreider2004unified} rely on synonymy within a shared ontology, an assumption that often breaks down in TCM--WM scenarios where correspondence depends on interpretation rather than lexical similarity~\cite{konopasky2020understanding}.

In LLM-based medical systems, inconsistencies in cross-system mappings can propagate through retrieval pipelines and result in unreliable evidence for downstream reasoning~\cite{sutton2020overview}.
This observation motivates alignment methods that jointly incorporate structural and semantic information while remaining compatible with query-driven retrieval.

Overall, existing approaches exhibit three main limitations: 
(1) alignment is typically formulated as context-independent pairwise matching, (2) retrieval models capture query relevance but do not model cross-system correspondence, and (3) medical knowledge integration relies on static mappings that are not adaptive to query semantics.
To address these limitations, we introduce QCEA, a query-conditioned, graph-aware, and direction-sensitive alignment framework.

\section{Proposed QCEA Framework}\label{sec:method}
\begin{figure*}
    \centering
    \includegraphics[width=\textwidth,keepaspectratio]{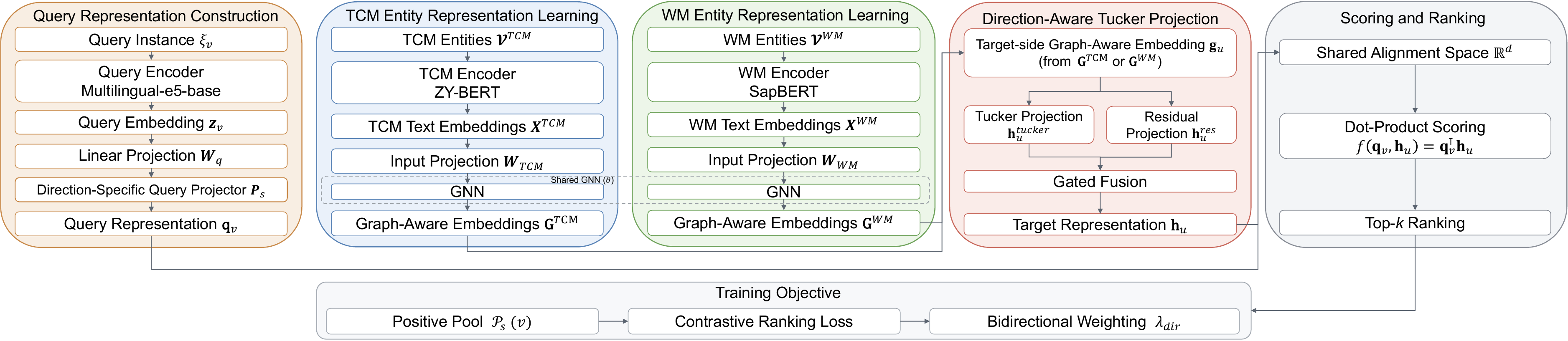}
    \caption{Overview of the proposed QCEA framework.
    Query descriptions are encoded into query representations.
    TCM and WM entities are mapped into graph-aware embeddings.
    A direction-aware Tucker projection module, conditioned on the alignment direction $s$, produces target representations for scoring and top-$k$ ranking under a direction-weighted, many-to-many-aware contrastive objective.}
    \label{fig:qcea_framework}
\end{figure*}
This section introduces QCEA, a query-conditioned alignment framework for cross-system medical knowledge integration.
Rather than learning a fixed correspondence function between entity pairs, QCEA constructs a query representation from the source-side entity description, learns graph-aware embeddings for TCM and WM entities, and projects target-side representations into a shared alignment space for ranking candidate entities.
An overview of the framework is shown in Fig.~\ref{fig:qcea_framework}.

\paragraph*{What is alignment?}
Alignment is not merely ranking relevance, but identifying \emph{cross-system semantic correspondence under representational mismatch}.
Unlike retrieval, where relevance is defined within a single semantic space, alignment requires establishing semantic correspondence between two heterogeneous spaces, where correspondence may be asymmetric, context-dependent, and non-bijective.
This distinction is fundamental: retrieval operates in a shared semantic space, while alignment must bridge two distinct representational systems.

\subsection{Problem Formulation}
We consider two knowledge graphs, $\mathcal{G}_{\text{TCM}}=(\mathcal{V}_{\text{TCM}},\mathcal{E}_{\text{TCM}})$ and $\mathcal{G}_{\text{WM}}=(\mathcal{V}_{\text{WM}},\mathcal{E}_{\text{WM}})$, representing TCM and WM systems, respectively.
Each entity $v$ is associated with a textual description $d_v$ composed of its name and definition.

A set of anchor correspondences $\mathcal{A}\subseteq \mathcal{V}_{\text{TCM}}\times\mathcal{V}_{\text{WM}}$ is available for supervision, where entities may have multiple counterparts in the opposite graph.

We define an alignment direction indicator $s \in \{0,1\}$, where $s=0$ denotes TCM$\rightarrow$WM alignment and $s=1$ denotes WM$\rightarrow$TCM alignment.
Given direction $s$, the source and target entity sets are defined as:
\begin{equation}
(\mathcal{V}_{\text{src}}^{(s)}, \mathcal{V}_{\text{tgt}}^{(s)}) =
\begin{cases}
(\mathcal{V}_{\text{TCM}}, \mathcal{V}_{\text{WM}}), & s=0, \\
(\mathcal{V}_{\text{WM}}, \mathcal{V}_{\text{TCM}}), & s=1.
\end{cases}
\end{equation}

Given a source entity $v \in \mathcal{V}_{\text{src}}^{(s)}$ and its textual description $d_v$, QCEA defines a query instance $\xi_v=(v,d_v)$ and derives a query representation $\mathbf{q}_v$ to rank candidate entities $u \in \mathcal{V}_{\text{tgt}}^{(s)}$:
\begin{equation}
\hat{u}=\arg\max_{u \in \mathcal{V}_{\text{tgt}}^{(s)}} f(\mathbf{q}_v,\mathbf{h}_u),
\end{equation}
where $\mathbf{h}_u$ denotes the target entity representation in the shared alignment space, defined by the target-side projection module below, and $f(\cdot,\cdot)$ denotes the alignment scoring function, instantiated as dot product in this work.

In practice, ranking may be performed either over the full target graph or under type constraints, depending on the evaluation setting.
This formulation recasts cross-system entity alignment as a context-dependent ranking problem, rather than static pairwise matching.
Unlike standard retrieval, the objective is not generic relevance estimation but alignment under asymmetric and potentially many-to-many correspondence, where a source entity may correspond to multiple target entities depending on direction and descriptive context.

\subsection{Framework Overview}
As illustrated in Fig.~\ref{fig:qcea_framework}, QCEA comprises five components: query representation construction, TCM entity representation learning, WM entity representation learning, direction-aware Tucker projection, and scoring/ranking.
These components are jointly optimized under a contrastive ranking objective that accounts for many-to-many correspondence through multi-positive sampling and bidirectional weighting.

\subsection{Query Representation Construction}
QCEA derives a query representation for each source entity from its textual description.
The query instance $\xi_v=(v,d_v)$ is encoded using multilingual-e5-base~\cite{wang2024multilingual} to obtain a dense embedding $\mathbf{z}_v\in\mathbb{R}^{d_q}$.
This query encoder captures fine-grained, context-dependent semantics from textual descriptions, which differ from entity representations optimized for structural consistency within each graph.
The encoder is kept fixed, and a subsequent projection layer adapts the representation to the alignment space:
\begin{equation}
\hat{\mathbf{z}}_v=\mathbf{W}_q\mathbf{z}_v,
\end{equation}
where $\mathbf{W}_q \in \mathbb{R}^{d \times d_q}$ is a trainable linear transformation.

This design relies on lightweight projection layers on top of a pretrained text encoder, rather than fine-tuning the encoder itself, thereby preserving general semantic structure captured during large-scale pretraining while adapting to the alignment task with limited supervision.
In contrast, directly fine-tuning the encoder under limited alignment supervision may lead to overfitting and reduced generalization.

The query embedding is mapped into the shared alignment space through a direction-specific projector:
\begin{equation}
\mathbf{q}_v=\mathrm{Norm}(\mathbf{P}_s\hat{\mathbf{z}}_v),
\end{equation}
where $\mathbf{P}_s \in \mathbb{R}^{d \times d}$ is a trainable projection matrix indexed by $s$, and $\mathrm{Norm}(\mathbf{x})=\mathbf{x}/\|\mathbf{x}\|_2$ denotes $\ell_2$ normalization.
The projector maps $\hat{\mathbf{z}}_v$ into a shared $d$-dimensional space (with $d=256$), consistent with the target-side representations.

\subsection{TCM and WM Entity Representation Learning}
QCEA initializes entity semantics on the two graph sides using domain-specific text encoders specialized for capturing entity-level semantics in their respective medical systems.
Specifically, TCM entities are encoded with ZY-BERT~\cite{mucheng2022tcm}, while WM entities are encoded with SapBERT~\cite{liu2021self}, both based on their names and definitions.
The resulting embeddings are precomputed and used as input features during training.
The text encoders are kept fixed to preserve general semantic structure, while subsequent modules are optimized on top of these representations to learn task-specific alignment patterns.

Let $\mathbf{x}^{\text{TCM}}_v\in\mathbb{R}^{d_T}$ and $\mathbf{x}^{\text{WM}}_u\in\mathbb{R}^{d_W}$ denote the initial text embeddings of TCM and WM entities.
As the two encoders operate in different semantic spaces, QCEA projects them into a shared node feature space:
\begin{equation}
\begin{aligned}
\tilde{\mathbf{x}}^{\text{TCM}}_v
&=
\mathrm{Norm}(\mathbf{W}_{\text{TCM}}\mathbf{x}^{\text{TCM}}_v),\\
\tilde{\mathbf{x}}^{\text{WM}}_u
&=
\mathrm{Norm}(\mathbf{W}_{\text{WM}}\mathbf{x}^{\text{WM}}_u),
\end{aligned}
\end{equation}
where $\mathbf{W}_{\text{TCM}}$ and $\mathbf{W}_{\text{WM}}$ are trainable projection matrices for the two graph sides.

To incorporate structural context, QCEA applies a shared graph encoder after input projection.
Let $\tilde{\mathbf{X}}^{\text{TCM}}\in\mathbb{R}^{N_{\text{TCM}}\times d}$ and $\tilde{\mathbf{X}}^{\text{WM}}\in\mathbb{R}^{N_{\text{WM}}\times d}$ denote the projected node feature matrices of the two graphs.

A graph encoder with shared parameters is applied to both graphs, with message passing performed independently on each graph topology:
\begin{equation}
\mathbf{G}^{\text{TCM}}=
\mathrm{GNN}_{\theta}(\tilde{\mathbf{X}}^{\text{TCM}},\mathcal{E}_{\text{TCM}}),
\end{equation}
\begin{equation}
\mathbf{G}^{\text{WM}}=
\mathrm{GNN}_{\theta}(\tilde{\mathbf{X}}^{\text{WM}},\mathcal{E}_{\text{WM}}),
\end{equation}
where $\mathrm{GNN}_{\theta}$ denotes a graph encoder with shared parameters $\theta$.
The resulting matrices $\mathbf{G}^{\text{TCM}}$ and $\mathbf{G}^{\text{WM}}$ represent graph-aware node embeddings.

This design preserves side-specific semantics through separate input projections while introducing a shared structural inductive bias via the shared graph encoder.
The encoder is instantiated as a GCN~\cite{kipf2017semi} without cross-graph message passing.
The resulting node representations are denoted as $\mathbf{g}^{\text{TCM}}_v$ and $\mathbf{g}^{\text{WM}}_u$.

\subsection{Direction-Aware Tucker Projection}
\paragraph*{Why Tucker?}
While simpler alternatives such as independent linear projections per direction are possible, we adopt Tucker decomposition for three reasons:
\begin{itemize}
    \item \textbf{Parameter sharing across directions}: the factorized form couples transformations via shared factors, reducing overfitting under limited alignment supervision.
    \item \textbf{Structured low-rank interactions}: unlike linear projections, the decomposition captures multiplicative dependencies between latent factors, yielding a richer transformation family.
    \item \textbf{Asymmetry with efficiency}: conditioning on alignment direction enables direction-specific mappings, improving generalization under asymmetric correspondence patterns while maintaining parameter efficiency through low-rank structure.
\end{itemize}

To map target-side graph embeddings into the shared alignment space, QCEA adopts a direction-aware Tucker projection module~\cite{kolda2009tensor}, whose transformation is conditioned on the alignment direction $s$.
The projection is implemented via a factorized multi-linear transformation with low-rank decomposition, capturing dependencies between latent factors in the embedding space.
By conditioning the factorization on $s$, the model enables direction-sensitive transformations while maintaining parameter efficiency.

For a candidate target entity $u \in \mathcal{V}_{\text{tgt}}^{(s)}$, its graph-aware embedding is defined as:
\begin{equation}
\mathbf{g}_u =
\begin{cases}
\mathbf{g}^{\text{WM}}_u, & s=0, \\
\mathbf{g}^{\text{TCM}}_u, & s=1.
\end{cases}
\end{equation}

Specifically, the transformation is defined by a direction factor matrix $\mathbf{U}_s\in\mathbb{R}^{2\times R_s}$, an output factor matrix $\mathbf{U}_o\in\mathbb{R}^{d\times R_o}$, an input factor matrix $\mathbf{U}_i \in \mathbb{R}^{d_{\text{in}} \times R_i}$, and a set of core slices $\{\mathbf{G}_r\}_{r=1}^{R_s}$ with $\mathbf{G}_r\in\mathbb{R}^{R_o\times R_i}$, where $d_{\text{in}}$ and $d$ denote the input and alignment embedding dimensions, respectively. 
In our formulation, $d_{\text{in}}=d$, since the graph-aware embeddings are already represented in the shared $d$-dimensional space.

Conditioned on direction $s$, the transformation is given by:
\begin{equation}
\mathbf{h}^{\text{tucker}}_u = \sum_{r=1}^{R_s} \mathbf{U}_s[s,r] \cdot \mathbf{U}_o \mathbf{G}_r \left(\mathbf{U}_i^\top \mathbf{g}_u\right),
\end{equation}
where $\mathbf{U}_s[s,r]$ denotes the $(s,r)$-th entry of $\mathbf{U}_s$.

Equivalently, the above transformation can be expressed in matrix form as:
\begin{equation}
\mathbf{W}^{(s)}
=
\mathbf{U}_o
\Big(\sum_{r=1}^{R_s} \mathbf{U}_s[s,r]\mathbf{G}_r\Big)
\mathbf{U}_i^\top.
\end{equation}
The transformed representation is:
\begin{equation}
\mathbf{h}^{\text{tucker}}_u=\mathbf{W}^{(s)}\mathbf{g}_u.
\end{equation}

To improve stability, we introduce a residual projection branch:
\begin{equation}
\mathbf{h}^{\text{res}}_u=\mathbf{R}\mathbf{g}_u,
\end{equation}
and combine it with the Tucker branch via a learnable scalar gate:
\begin{equation}
\mathbf{h}_u=
(1-\sigma(\alpha))\mathbf{h}^{\text{tucker}}_u
+
\sigma(\alpha)\mathbf{h}^{\text{res}}_u,
\end{equation}
followed by $\ell_2$ normalization. 
The residual branch provides a direct linear mapping, while the gate controls the trade-off between the Tucker-based transformation and the residual pathway.

\subsection{Scoring and Ranking}

After projection, the query representation $\mathbf{q}_v$ and target representation $\mathbf{h}_u$ are compared in the shared alignment space via dot product:
\begin{equation}
f(\mathbf{q}_v,\mathbf{h}_u)=\mathbf{q}_v^\top\mathbf{h}_u.
\end{equation}
Since both representations are $\ell_2$-normalized, the dot product is equivalent to cosine similarity.
Scores are used to rank candidate entities and obtain top-$k$ predictions.

At inference time, ranking is performed over a candidate set of target entities under the specified alignment direction.
Depending on the evaluation setting, this candidate set may be the entire target graph or a subset restricted to the same entity type.
For each query entity $v$, candidate entities are scored and sorted according to $f(\mathbf{q}_v,\mathbf{h}_u)$, and the top-$k$ ranked entities are returned as alignment predictions.

\subsection{Many-to-Many Contrastive Ranking Objective}

Cross-system medical alignment is inherently many-to-many, where a source entity may correspond to multiple valid targets.
Accordingly, QCEA is trained with a contrastive ranking objective that supports multiple positives per query, rather than a single-positive classification loss.

For each source entity $v \in \mathcal{V}_{\text{src}}^{(s)}$, we define the positive pool under direction $s$ as:
\begin{equation}
\mathcal{P}_s(v)=\{u \in \mathcal{V}_{\text{tgt}}^{(s)} \mid (v,u)\in \mathcal{A}^{(s)}\},
\end{equation}
where $\mathcal{A}^{(0)}=\mathcal{A}$ and $\mathcal{A}^{(1)}=\{(u,v)\mid (v,u)\in\mathcal{A}\}$ denotes the reversed correspondence set under the opposite alignment direction.

During training, we adopt a stochastic sampling strategy to construct training instances.
Specifically, we first obtain a training-time positive pool $\mathcal{P}_s^{\text{train}}(v)$, which contains only those correspondences in $\mathcal{P}_s(v)$ that are observed in the training split.
Thus, $\mathcal{P}_s^{\text{train}}(v) \subseteq \mathcal{P}_s(v)$.
Up to $P$ positives are then sampled uniformly without replacement from this pool.
To ensure training stability, the current ground-truth target associated with the query is always included in the sampled positives.

Negative samples are drawn uniformly from the entire target graph, excluding all known positives in the global pool $\mathcal{P}_s(v)$.
This avoids false negatives from valid but unobserved correspondences.

Let $\{u^+_1,\ldots,u^+_P\}\subseteq\mathcal{P}_s^{\text{train}}(v)$ denote sampled positives and $\{u^-_1,\ldots,u^-_K\}$ denote sampled negatives. 
The corresponding logits are:
\begin{equation}
\ell_i^+=f(\mathbf{q}_v,\mathbf{h}_{u_i^+}),\quad
\ell_j^-=f(\mathbf{q}_v,\mathbf{h}_{u_j^-}).
\end{equation}

QCEA optimizes a multi-positive contrastive objective related to InfoNCE~\cite{oord2018representation} and supervised contrastive learning~\cite{khosla2020supervised}:
\begin{equation}
\mathcal{L}_{\text{mp}}
=
-\log
\frac{\sum_{i=1}^{P} \exp(\ell_i^+/\tau)}
{\sum_{i=1}^{P} \exp(\ell_i^+/\tau)+\sum_{j=1}^{K} \exp(\ell_j^-/\tau)},
\end{equation}
where $\tau$ is a temperature parameter.

Alignment is learned in both TCM$\rightarrow$WM and WM$\rightarrow$TCM directions. 
The overall objective is defined as:
\begin{equation}
\mathcal{L}
=
\lambda_{\text{dir}}\mathcal{L}_{\text{WM}\rightarrow\text{TCM}}
+
(1-\lambda_{\text{dir}})\mathcal{L}_{\text{TCM}\rightarrow\text{WM}}
+
\lambda_{\text{reg}}\mathcal{L}_{\text{reg}},
\end{equation}
where $\lambda_{\text{dir}}$ balances the two alignment directions and $\lambda_{\text{reg}}$ controls the $\ell_2$ regularization term $\mathcal{L}_{\text{reg}}$.

\subsection{Discussion}
QCEA departs from conventional entity alignment in three aspects.
It replaces static pairwise matching with query-conditioned ranking, decouples query construction from graph-aware entity representation learning, and introduces a direction-aware Tucker projection with residual fusion for target-side transformation.

Under this formulation, bidirectional training induces a many-to-many alignment structure, which is naturally optimized via a multi-positive contrastive objective.
In contrast to retrieval models and LLM-enhanced alignment methods that focus on relevance estimation or pairwise similarity, QCEA formulates alignment as a query-conditioned ranking problem, aligning the training objective with retrieval-based inference.

\paragraph*{Theoretical insight.}
When cross-system correspondence is context-dependent, a fixed deterministic correspondence between entities is inherently insufficient, as it cannot represent one-to-many or description-dependent alignments within a single function.
In such cases, the alignment target becomes conditional on the query context rather than solely determined by the source entity, since the same source concept may correspond to different targets under different descriptions.
QCEA addresses this limitation by modeling alignment as a query-conditioned ranking function, enabling context-aware and non-bijective correspondence.

\section{Experimental Setup}
We evaluate QCEA at both the alignment level and the system level, assessing its effectiveness for cross-system retrieval and reasoning in LLM-based medical applications.

\subsection{Datasets}
We construct a cross-system medical entity alignment benchmark between TCM and WM, consisting of two tasks: 
\textbf{(1) symptom alignment} (TCM symptom $\leftrightarrow$ WM symptom) and 
\textbf{(2) herb--molecule alignment} (TCM herb $\leftrightarrow$ WM molecule).

The benchmark is derived from SymMap~\cite{wu2019symmap}, a publicly available resource linking TCM and biomedical entities.
To ensure clinical relevance, we adopt an ICD-guided, disease-centric construction strategy by selecting respiratory-related diseases (ICD-10-CM J00--J99) and extracting subgraphs of associated symptoms, TCM herbs, and WM molecules.
ICD codes are used for domain scoping rather than defining alignment supervision. 
This construction yields clinically grounded associations and naturally introduces many-to-many and asymmetric correspondences.

The resulting TCM and WM graphs contain 1,048 and 3,568 entities, with 6,012 and 13,176 edges, respectively. 
After deduplication, the benchmark includes 790 symptom correspondences and 19,440 herb--molecule correspondences, with multiple associations per entity.

The two tasks exhibit distinct alignment characteristics. 
Symptom alignment is near one-to-one, with most entities associated with a small number of counterparts. 
In contrast, herb--molecule alignment exhibits substantially higher cardinality, with each TCM herb linked to more than 20 WM molecules on average, and some exceeding 100. 

Each entity is associated with textual descriptions (name and definition), which are used to construct query representations. 
The benchmark is clinically grounded yet domain-scoped, providing a controlled setting for studying cross-system alignment under semantic heterogeneity, rather than targeting broad generalization across all disease domains.

This disease-centric construction retains approximately 18.6\% of the original SymMap connectivity and may introduce bias toward well-documented respiratory conditions.

\subsection{Evaluation Protocol}

Entity alignment is evaluated as a ranking task, where a query derived from a source entity ranks candidates in the target graph.
We consider two retrieval settings: 
\textbf{(1) type-constrained retrieval}, which restricts candidates to semantically compatible categories, and 
\textbf{(2) full retrieval}, which considers all entities in the target graph.
We adopt a multi-positive, group-level evaluation protocol and report Hit@K, Recall@K, and MRR.
Hit@K measures whether at least one correct entity appears in the top-$k$ predictions, Recall@K measures the proportion of ground-truth entities retrieved within the top-$k$, and MRR evaluates the average reciprocal rank of the highest-ranked correct entity.
For the Herb task, Recall@100 is additionally reported due to the larger candidate space.
Among these metrics, we consider Hit@10 and MRR as the primary evaluation criteria, as they reflect top-rank performance most relevant to downstream retrieval tasks.
Hit@1 is reported as a secondary indicator of precision, while Recall@K provides supplementary coverage analysis.

The dataset is split into training, validation, and test sets with a 60\%/20\%/20\% ratio at the alignment-pair level.
Due to many-to-many correspondence, the same source entity may appear in different splits with different target entities.

During evaluation, we adopt grouped multi-positive ranking, where all ground-truth targets for each query are evaluated jointly as a set.
Model selection is based on validation Hit@10 with early stopping, and performance is reported on the test set.
We further report stratified results by alignment direction (TCM$\rightarrow$WM vs.\ WM$\rightarrow$TCM) and ground-truth cardinality (GT$=$1 vs.\ GT$>$1).

\paragraph*{Remark on data splitting.}
Although the same entity may appear across training, validation, and test splits, alignment supervision (i.e., entity pairs) is strictly separated.
This setting evaluates generalization to unseen correspondences rather than unseen entities, which is consistent with transductive graph learning scenarios where entity vocabularies are fixed but alignment patterns vary.

\subsection{Implementation Details}

Entities from the TCM and WM graphs are encoded using domain-specific pretrained models and projected into a shared space with dimension $d=256$.
A shared GCN is used to incorporate structural information.

The direction-aware Tucker projection adopts rank $(R_s, R_o, R_i) = (16, 128, 128)$.
Models are trained with a many-to-many contrastive ranking objective using multi-positive sampling and bidirectional weighting.
The direction weight $\lambda_{\text{dir}}$ is set to 0.5 for Symptom and 0.3 for Herb, reflecting differences in alignment asymmetry across tasks.

We report results averaged over five random seeds, with standard deviation below 0.01 across main metrics, indicating stable training behavior.
Training is conducted for 300 epochs using Adam~\cite{kingma2014adam} with a learning rate of $1\times10^{-3}$, batch size 64, and 1,024 negative samples per query, chosen to balance performance and computational cost.
We additionally employ early stopping, gradient clipping, and learning-rate decay to improve optimization stability.

\subsection{Compared Methods}
We compare QCEA with representative baselines covering different alignment paradigms:
\textbf{(1) Text-only Bi-Encoder}, a text retrieval baseline that matches query and entity embeddings via cosine similarity using separate encoders and the same textual inputs as QCEA, without graph or direction-aware modeling, thereby isolating the effect of alignment modeling;
\textbf{(2) Cross-Attention}, implemented as a lightweight cross-encoder with multi-head attention over projected token-level representations and a learned scoring function, modeling bidirectional interactions between query and candidate embeddings;
\textbf{(3) MLP}, implemented as a parametric matching function over concatenated query–entity embeddings;
\textbf{(4) Procrustes}~\cite{conneau2018word}, learning a linear mapping with nearest-neighbor alignment;
\textbf{(5) GCN-Align}~\cite{wang2018cross}, capturing structural consistency via graph convolution;
\textbf{(6) RDGCN}~\cite{wu2019relation}, extending GCN-based alignment with relation-aware interactions.

All neural baselines operate on precomputed text embeddings without additional encoder fine-tuning and are trained with the same multi-positive and negative sampling strategy.
These baselines cover semantic, interaction-based, parametric, geometric, and structural alignment paradigms.

All methods are evaluated under the same candidate sets following a unified protocol, including multi-positive ranking and group-based metrics, under both \textbf{type-constrained} and \textbf{full-retrieval} settings.

We further include ablation variants of QCEA by removing key components, including query conditioning, shared graph propagation, direction-aware Tucker projection, and residual fusion.

\paragraph*{Comparison with LLM-Based Alignment Methods.}
LLM-based alignment methods (e.g., prompting GPT-4~\cite{achiam2023gpt} for free-form entity matching) operate in an open-ended generation setting without a predefined candidate space, whereas our evaluation focuses on ranking within a fixed candidate set.
This reflects a fundamental difference in problem formulation between generation-based matching and candidate ranking, making direct comparison infeasible.
Exploring hybrid approaches that combine dense retrieval with LLM-based re-ranking remains an interesting direction for future work.

\subsection{Downstream RAG Evaluation}

To assess whether improvements in cross-system alignment translate to downstream performance, we construct a controlled evaluation under a unified RAG framework.
All RAG-based settings share the same QA benchmark, generation model, decoding configuration, prompt template, and answer normalization rules, isolating the effect of alignment quality.
The QA benchmark is constructed solely for downstream evaluation and is not used during model training.
Answer annotations are separated from alignment supervision, and the generation model does not access gold answers during training or inference.

To facilitate reproducibility, we will release the benchmark construction protocol, evaluation scripts, and prompt templates upon acceptance.
The benchmark contains 400 questions across eight categories, including four single-hop and four two-hop tasks.
Single-hop tasks evaluate direct cross-system transfer, while two-hop tasks require additional intra-graph reasoning following cross-system transitions.
Each question is associated with reference answers, gold evidence, and validation rules, enabling consistent evaluation of retrieval, correctness, and grounding.

We compare six settings: 
\textbf{Oracle} (using ground-truth alignments),
\textbf{QCEA} (using predicted first-hop alignment candidates),
\textbf{QCEA-TopX} (retaining only the top-$X$ ranked candidates to control alignment precision, where $X$ is varied from 1 to 10),
\textbf{QCEA-DropX} (randomly removing a subset of candidates to test robustness to candidate loss at different removal ratios),
\textbf{NoAlign} (removing cross-system links), 
and \textbf{Only-LLM} (no retrieval).

All RAG-based settings use the same generation model (Qwen2.5-1.5B-Instruct~\cite{yang2024qwen25}), ensuring a controlled comparison across alignment settings.

For the Only-LLM setting, we additionally evaluate multiple language models with different capability profiles, including a larger open general-purpose model (gpt-oss-20B~\cite{agarwal2025gpt}), a TCM-oriented model (CMLM-ZhongJing~\cite{yang2024zhongjing}), and a WM-oriented reasoning model (MedReason-8B~\cite{wu2025medreason}), to examine whether increased model capacity or domain specialization can compensate for the absence of cross-system alignment.

This design isolates the effect of first-hop alignment quality and evidence availability on downstream medical question answering.

\paragraph*{Choice of Language Model.}
We intentionally adopt a lightweight language model to better expose the impact of alignment quality on downstream reasoning, as smaller models are less capable of compensating for missing or incorrect evidence.
This design choice isolates the effect of alignment from the model's parametric knowledge, providing a clearer view of how alignment quality influences retrieval and grounding.
While larger models may exhibit different trade-offs, our focus is on understanding the relative impact of alignment under controlled conditions rather than achieving maximum absolute performance.

We evaluate performance at three levels.
\textbf{Retrieval-level} metrics include \emph{evidence recall@K} and \emph{cross-system hit rate}, measuring evidence coverage and cross-system alignment success.
\textbf{Generation-level} metrics include \emph{answer accuracy}, \emph{answer accuracy (strict)}, and \emph{two-hop slot accuracy}.
\textbf{End-to-end} metrics include \emph{groundedness}, \emph{end-to-end accuracy}, \emph{end-to-end accuracy (strict)}, and \emph{hallucination rate}.

For Only-LLM, retrieval metrics are not applicable.
In this case, without retrieval, end-to-end accuracy reduces to \emph{answer accuracy}, and strict end-to-end accuracy reduces to \emph{answer accuracy (strict)}.

\section{Experiment Results}
\subsection{Comparison with Baseline Methods}
\begin{table*}
\centering
\caption{Comparison with baseline methods under \textbf{type-constrained} and \textbf{full-retrieval} settings. Recall@100 is used for Herb. The best and second-best results in each column are highlighted in \textbf{bold} and underlined, respectively.}
\label{tab:main_results_dual}

\setlength{\tabcolsep}{1.8pt}
\renewcommand{\arraystretch}{0.9}
\scriptsize

\resizebox{\textwidth}{!}{
\begin{tabular}{lcccccccccccccccc}
\toprule
& \multicolumn{8}{c}{\textbf{Symptom}} & \multicolumn{8}{c}{\textbf{Herb}} \\
\cmidrule(lr){2-9}\cmidrule(lr){10-17}
& \multicolumn{4}{c}{\textbf{Full}} & \multicolumn{4}{c}{\textbf{Type}} 
& \multicolumn{4}{c}{\textbf{Full}} & \multicolumn{4}{c}{\textbf{Type}} \\
\cmidrule(lr){2-5}\cmidrule(lr){6-9}\cmidrule(lr){10-13}\cmidrule(lr){14-17}
\textbf{Method}
& Hit@1 & Hit@10 & Recall@10 & MRR
& Hit@1 & Hit@10 & Recall@10 & MRR
& Hit@1 & Hit@10 & Recall@100 & MRR
& Hit@1 & Hit@10 & Recall@100 & MRR \\
\midrule

GCN-Align 
& \underline{0.4609} & 0.8696 & 0.7986 & \underline{0.6000}
& \underline{0.4609} & 0.8783 & 0.8072 & \underline{0.6015}
& 0.0573 & 0.3512 & 0.4448 & 0.1449
& 0.0573 & 0.3512 & 0.4448 & 0.1449 \\

RDGCN
& 0.4000 & 0.8174 & 0.8174 & 0.5305
& 0.4000 & 0.8174 & 0.8174 & 0.5313
& 0.0845 & \underline{0.3962} & \textbf{0.7653} & 0.1835
& 0.0845 & \underline{0.3962} & \textbf{0.7662} & 0.1835 \\

MLP 
& 0.0261 & 0.2696 & 0.2283 & 0.1135
& 0.0261 & 0.3043 & 0.2630 & 0.1271
& 0.0808 & 0.3700 & 0.5319 & 0.1720
& 0.0808 & 0.3700 & 0.5319 & 0.1720 \\

Text-only Bi-Encoder
& 0.1652 & 0.5913 & 0.5605 & 0.2971
& 0.1739 & 0.6957 & 0.6612 & 0.3240
& 0.0019 & 0.0113 & 0.0169 & 0.0069
& 0.0075 & 0.0507 & 0.1587 & 0.0288 \\

Cross-Attention
& 0.0000 & 0.0087 & 0.0029 & 0.0066
& 0.0870 & 0.3652 & 0.3536 & 0.1679
& 0.0032 & 0.0374 & 0.1295 & 0.0198
& 0.0032 & 0.0438 & 0.1466 & 0.0241 \\

Procrustes
& 0.4261 & \underline{0.8870} & \underline{0.8304} & 0.5854
& 0.4261 & \underline{0.8870} & \underline{0.8315} & 0.5861
& \textbf{0.1070} & 0.3446 & 0.4882 & \underline{0.1868}
& \textbf{0.1070} & 0.3446 & 0.4882 & \underline{0.1868} \\

QCEA 
& \textbf{0.5130} & \textbf{0.9130} & \textbf{0.8572} & \textbf{0.6293}
& \textbf{0.5130} & \textbf{0.9304} & \textbf{0.8746} & \textbf{0.6302}
& \underline{0.0958} & \textbf{0.4751} & \underline{0.5967} & \textbf{0.2087}
& \underline{0.0958} & \textbf{0.4751} & \underline{0.6069} & \textbf{0.2094} \\

\midrule
\multicolumn{17}{c}{\textbf{TCM$\rightarrow$WM}} \\
\midrule

GCN-Align 
& \underline{0.6250} & \textbf{0.9750} & \textbf{0.9750} & \underline{0.7558}
& \underline{0.6250} & \textbf{0.9750} & \textbf{0.9750} & \underline{0.7568}
& \underline{0.1026} & \textbf{0.5684} & \underline{0.5393} & 0.2382
& \underline{0.1026} & \textbf{0.5684} & \underline{0.5393} & 0.2382 \\

RDGCN
& 0.5750 & 0.8875 & 0.8875 & 0.6884
& 0.5750 & 0.8875 & 0.8875 & 0.6893
& \textbf{0.1132} & 0.5316 & \textbf{0.8342} & \underline{0.2420}
& \textbf{0.1132} & 0.5316 & \textbf{0.8342} & \underline{0.2420} \\

MLP
& 0.0000 & 0.2500 & 0.2500 & 0.0935
& 0.0000 & 0.3000 & 0.3000 & 0.1128
& \textbf{0.1132} & 0.5368 & 0.5283 & \textbf{0.2470}
& \textbf{0.1132} & 0.5368 & 0.5283 & \textbf{0.2470} \\

Text-only Bi-Encoder
& 0.2125 & 0.6875 & 0.6875 & 0.3552
& 0.2125 & 0.7875 & 0.7875 & 0.3705
& 0.0053 & 0.0316 & 0.0215 & 0.0137
& 0.0053 & 0.0342 & 0.0303 & 0.0182 \\

Cross-Attention
& 0.0000 & 0.0000 & 0.0000 & 0.0011
& 0.1125 & 0.5000 & 0.5000 & 0.2199
& 0.0000 & 0.0108 & 0.0318 & 0.0084
& 0.0000 & 0.0189 & 0.0443 & 0.0137 \\

Procrustes
& 0.5125 & 0.8625 & 0.8625 & 0.6319
& 0.5125 & 0.8625 & 0.8625 & 0.6321
& 0.0211 & 0.1579 & 0.2038 & 0.0702
& 0.0211 & 0.1579 & 0.2038 & 0.0702 \\

QCEA 
& \textbf{0.6750} & \underline{0.9625} & \underline{0.9625} & \textbf{0.7663}
& \textbf{0.6750} & \underline{0.9625} & \underline{0.9625} & \textbf{0.7664}
& 0.1000 & \underline{0.5579} & \underline{0.5647} & 0.2274
& 0.1000 & \underline{0.5579} & \underline{0.5647} & 0.2274 \\

\midrule
\multicolumn{17}{c}{\textbf{WM$\rightarrow$TCM}} \\
\midrule

GCN-Align 
& 0.0857 & 0.6286 & 0.3952 & 0.2439
& 0.0857 & 0.6571 & 0.4238 & 0.2465
& 0.0321 & 0.2307 & 0.3924 & 0.0932
& 0.0321 & 0.2307 & 0.3924 & 0.0932 \\

RDGCN
& 0.0000 & \underline{0.6571} & \underline{0.6571} & 0.1697
& 0.0000 & 0.6571 & 0.6571 & 0.1701
& 0.0686 & 0.3212 & \textbf{0.7270} & 0.1510
& 0.0686 & 0.3212 & \textbf{0.7285} & 0.1510 \\

MLP
& 0.0857 & 0.3143 & 0.1786 & 0.1591
& 0.0857 & 0.3143 & 0.1786 & 0.1598
& 0.0628 & 0.2774 & 0.5339 & 0.1304
& 0.0628 & 0.2774 & 0.5339 & 0.1304 \\

Text-only Bi-Encoder
& 0.0571 & 0.3714 & 0.2702 & 0.1641
& 0.0857 & 0.4857 & 0.3726 & 0.2178
& 0.0000 & 0.0000 & 0.0143 & 0.0031
& 0.0088 & 0.0599 & 0.2300 & 0.0347 \\

Cross-Attention
& 0.0000 & 0.0286 & 0.0095 & 0.0193
& 0.0286 & 0.0571 & 0.0190 & 0.0490
& 0.0053 & 0.0548 & 0.1933 & 0.0272
& 0.0053 & 0.0601 & 0.2135 & 0.0309 \\

Procrustes
& \textbf{0.2286} & \textbf{0.9429} & \textbf{0.7571} & \textbf{0.4791}
& \textbf{0.2286} & \textbf{0.9429} & \textbf{0.7607} & \textbf{0.4808}
& \textbf{0.1547} & \textbf{0.4482} & \underline{0.6460} & \textbf{0.2515}
& \textbf{0.1547} & \textbf{0.4482} & \underline{0.6460} & \textbf{0.2515} \\

QCEA 
& \underline{0.1429} & \underline{0.8000} & 0.6167 & \underline{0.3161}
& \underline{0.1429} & \underline{0.8571} & \underline{0.6738} & \underline{0.3189}
& \underline{0.0934} & \underline{0.4292} & 0.6145 & \underline{0.1983}
& \underline{0.0934} & \underline{0.4292} & 0.6302 & \underline{0.1994} \\

\midrule
\multicolumn{17}{c}{\textbf{GT$=$1}} \\
\midrule

GCN-Align 
& \underline{0.5258} & \underline{0.8866} & \underline{0.8866} & \underline{0.6528}
& \underline{0.5258} & \underline{0.8969} & \underline{0.8969} & \underline{0.6545}
& 0.0150 & 0.1175 & 0.4188 & 0.0538
& 0.0150 & 0.1175 & 0.4188 & 0.0538 \\

RDGCN
& 0.4742 & 0.8351 & 0.8351 & 0.5973
& 0.4742 & 0.8351 & 0.8351 & 0.5981
& 0.0641 & 0.2137 & \underline{0.6175} & 0.1174
& 0.0641 & 0.2137 & \underline{0.6197} & 0.1175 \\

MLP
& 0.0000 & 0.2165 & 0.2165 & 0.0802
& 0.0000 & 0.2577 & 0.2577 & 0.0963
& 0.0406 & 0.1688 & 0.5470 & 0.0846
& 0.0406 & 0.1688 & 0.5470 & 0.0846 \\

Text-only Bi-Encoder
& 0.1753 & 0.6082 & 0.6082 & 0.3092
& 0.1753 & 0.7113 & 0.7113 & 0.3269
& 0.0000 & 0.0021 & 0.0192 & 0.0027
& 0.0043 & 0.0321 & 0.2179 & 0.0195 \\

Cross-Attention
& 0.0000 & 0.0000 & 0.0000 & 0.0017
& 0.0928 & 0.4124 & 0.4124 & 0.1835
& 0.0028 & 0.0168 & 0.1592 & 0.0124
& 0.0028 & 0.0168 & 0.1676 & 0.0136 \\

Procrustes
& 0.4536 & 0.8660 & 0.8660 & 0.5983
& 0.4536 & 0.8660 & 0.8660 & 0.5991
& \textbf{0.1068} & \textbf{0.3162} & \textbf{0.6496} & \textbf{0.1748}
& \textbf{0.1068} & \textbf{0.3162} & \textbf{0.6496} & \textbf{0.1748} \\

QCEA 
& \textbf{0.5876} & \textbf{0.9072} & \textbf{0.9072} & \textbf{0.6872}
& \textbf{0.5876} & \textbf{0.9278} & \textbf{0.9278} & \textbf{0.6883}
& \underline{0.0705} & \underline{0.2457} & 0.5641 & \underline{0.1324}
& \underline{0.0705} & \underline{0.2457} & 0.5812 & \underline{0.1334} \\

\midrule
\multicolumn{17}{c}{\textbf{GT$>$1}} \\
\midrule

GCN-Align 
& 0.1111 & 0.7778 & 0.3241 & 0.3155
& 0.1111 & 0.7778 & 0.3241 & 0.3159
& 0.0905 & 0.5343 & 0.4652 & 0.2164
& 0.0905 & 0.5343 & 0.4652 & 0.2164 \\

RDGCN
& 0.0000 & 0.7222 & \textbf{0.7222} & 0.1708
& 0.0000 & 0.7222 & \textbf{0.7222} & 0.1713
& 0.1005 & \underline{0.5394} & \textbf{0.8811} & 0.2352
& 0.1005 & \underline{0.5394} & \textbf{0.8811} & 0.2352 \\

MLP
& \underline{0.1667} & 0.5556 & 0.2917 & \underline{0.2927}
& \underline{0.1667} & 0.5556 & 0.2917 & \underline{0.2929}
& \underline{0.1122} & 0.5276 & 0.5200 & \underline{0.2405}
& \underline{0.1122} & 0.5276 & 0.5200 & \underline{0.2405} \\

Text-only Bi-Encoder
& 0.1111 & 0.5000 & 0.3032 & 0.2315
& \underline{0.1667} & 0.6111 & 0.3912 & 0.3086
& 0.0034 & 0.0184 & 0.0150 & 0.0102
& 0.0101 & 0.0653 & 0.1123 & 0.0361 \\

Cross-Attention
& 0.0000 & 0.0556 & 0.0185 & 0.0335
& 0.0556 & 0.1111 & 0.0370 & 0.0838
& 0.0035 & 0.0502 & 0.1111 & 0.0243
& 0.0035 & 0.0606 & 0.1336 & 0.0307 \\

Procrustes
& \textbf{0.2778} & \textbf{1.0000} & \underline{0.6389} & \textbf{0.5159}
& \textbf{0.2778} & \textbf{1.0000} & \underline{0.6458} & \textbf{0.5159}
& 0.1072 & 0.3668 & 0.3618 & 0.1962
& 0.1072 & 0.3668 & 0.3618 & 0.1962 \\

QCEA 
& 0.1111 & \underline{0.9444} & 0.5880 & \underline{0.3172}
& 0.1111 & \underline{0.9444} & 0.5880 & \underline{0.3172}
& \textbf{0.1156} & \textbf{0.6549} & \underline{0.6223} & \textbf{0.2685}
& \textbf{0.1156} & \textbf{0.6549} & \underline{0.6270} & \textbf{0.2689} \\

\bottomrule
\end{tabular}
}
\end{table*}

Table~\ref{tab:main_results_dual} summarizes the main alignment results.
The baselines represent structural methods (GCN-Align, RDGCN), geometric alignment (Procrustes), parametric matching (MLP), and semantic/interaction-based models (Text-only Bi-Encoder, Cross-Attention), which exhibit different performance characteristics across tasks and settings.

Overall, QCEA achieves the strongest and most consistent performance on the Symptom task.
Under both full-retrieval and type-constrained settings, it attains the top results across all Symptom metrics, indicating the effectiveness of query-conditioned ranking when cross-system correspondence is semantically nuanced but recoverable through contextualized matching.
Under full retrieval, QCEA reaches Hit@1/Hit@10/Recall@10/MRR of 0.5130/0.9130/0.8572/0.6293, and maintains similar advantages under type-constrained retrieval (0.5130/0.9304/0.8746/0.6302).

On Herb, performance is more differentiated.
QCEA achieves the best Hit@10 and MRR under both settings, while Procrustes attains the best Hit@1 and RDGCN the best Recall@100, reflecting a trade-off between top-rank precision and broader candidate coverage.
This finding suggests that QCEA concentrates relevant targets near the top of the ranking, whereas structurally oriented methods such as RDGCN remain advantageous when evaluation emphasizes wider recall; in practice, downstream retrieval and reasoning are typically more sensitive to top-ranked candidates than to deeper candidate coverage.

Baseline methods exhibit complementary strengths.
Graph-based methods remain competitive when mappings are relatively stable, with RDGCN performing particularly well in high-recall scenarios.
Procrustes performs well when correspondences are more regular or globally compressible (e.g., WM$\rightarrow$TCM Symptom).
In contrast, Text-only Bi-Encoder and Cross-Attention are less robust, especially on Herb, highlighting the limitation of relying solely on semantic interaction without structural modeling.

Directional and cardinality analyses further clarify where QCEA is most beneficial.
For Symptom, QCEA achieves the best Hit@1 and MRR in the TCM$\rightarrow$WM direction while remaining strongest overall.
For Herb with GT$>$1, QCEA achieves the best Hit@1, Hit@10, and MRR, whereas RDGCN remains strongest on Recall@100.
These results suggest that QCEA is particularly effective in ambiguous and multi-target scenarios, where improving the top of the ranking is more critical than expanding deeper candidate coverage.

At the same time, QCEA does not dominate all scenarios.
A plausible explanation is that WM$\rightarrow$TCM often involves mapping relatively standardized biomedical concepts to broader and more heterogeneous TCM concepts, making the reverse direction intrinsically more ambiguous and more sensitive to coarse global regularities.
GCN-Align remains stronger on TCM$\rightarrow$WM Symptom Hit@10/Recall@10, RDGCN on Herb Recall@100, and Procrustes on WM$\rightarrow$TCM Symptom, indicating that simpler structural or linear methods can remain effective when correspondence patterns are more standardized. 
Overall, the results support modeling cross-system entity alignment as a query-conditioned ranking problem, especially when evaluation emphasizes top-rank quality directly relevant to downstream retrieval and RAG.

\subsection{Ablation Study and Component Analysis}
\begin{table}
\centering
\caption{
Ablation results under \textbf{type-constrained} evaluation. Recall@100 is reported for Herb. Best results are in \textbf{bold}.
Variants: A (w/o query conditioning), B (full model), C (w/o GNN), D (linear projection), and E (w/o residual fusion).
}
\label{tab:ablation_unified}

\setlength{\tabcolsep}{2.2pt}
\renewcommand{\arraystretch}{0.88}
\scriptsize

\resizebox{\columnwidth}{!}{
\begin{tabular}{lcccccccc}
\toprule
& \multicolumn{4}{c}{\textbf{Symptom}} & \multicolumn{4}{c}{\textbf{Herb}} \\
\cmidrule(lr){2-5}\cmidrule(lr){6-9}
\textbf{Method} 
& \textbf{Hit@1} & \textbf{Hit@10} & \textbf{Recall@10} & \textbf{MRR}
& \textbf{Hit@1} & \textbf{Hit@10} & \textbf{Recall@100} & \textbf{MRR} \\
\midrule

\multicolumn{9}{c}{\textbf{Overall}} \\
\cmidrule(lr){1-9}
A & 0.3304 & 0.8000 & 0.7370 & 0.4911
  & 0.0423 & 0.3146 & 0.4578 & 0.1249 \\
B & \textbf{0.5130} & \textbf{0.9304} & \textbf{0.8746} & \textbf{0.6302}
  & \textbf{0.0958} & \textbf{0.4751} & \textbf{0.6069} & \textbf{0.2094} \\
C & 0.3217 & 0.7913 & 0.7453 & 0.4463
  & 0.0732 & 0.3531 & 0.5103 & 0.1617 \\
D & 0.3652 & 0.8261 & 0.7685 & 0.5302
  & 0.0648 & 0.3521 & 0.4847 & 0.1549 \\
E & 0.1043 & 0.5217 & 0.4696 & 0.2283
  & 0.0704 & 0.3596 & 0.5029 & 0.1621 \\

\midrule
\multicolumn{9}{c}{\textbf{TCM$\rightarrow$WM}} \\
\cmidrule(lr){1-9}
A & 0.4625 & 0.8625 & 0.8625 & 0.5971
  & 0.0579 & 0.4500 & 0.5196 & 0.1721 \\
B & \textbf{0.6750} & \textbf{0.9625} & \textbf{0.9625} & \textbf{0.7664}
  & 0.1000 & \textbf{0.5579} & \textbf{0.5647} & 0.2274 \\
C & 0.4625 & 0.8875 & 0.8875 & 0.5699
  & 0.1079 & 0.5053 & 0.5222 & 0.2343 \\
D & 0.5000 & 0.9125 & 0.9125 & 0.6526
  & 0.0974 & 0.5079 & 0.5522 & 0.2174 \\
E & 0.1250 & 0.5875 & 0.5875 & 0.2723
  & \textbf{0.1158} & 0.5368 & 0.5535 & \textbf{0.2478} \\

\midrule
\multicolumn{9}{c}{\textbf{WM$\rightarrow$TCM}} \\
\cmidrule(lr){1-9}
A & 0.0286 & 0.6571 & 0.4500 & 0.2488
  & 0.0336 & 0.2394 & 0.4235 & 0.0987 \\
B & \textbf{0.1429} & \textbf{0.8571} & \textbf{0.6738} & \textbf{0.3189}
  & \textbf{0.0934} & \textbf{0.4292} & \textbf{0.6302} & \textbf{0.1994} \\
C & 0.0000 & 0.5714 & 0.4202 & 0.1640
  & 0.0540 & 0.2686 & 0.5037 & 0.1214 \\
D & 0.0571 & 0.6286 & 0.4393 & 0.2505
  & 0.0467 & 0.2657 & 0.4473 & 0.1201 \\
E & 0.0571 & 0.3714 & 0.2000 & 0.1277
  & 0.0453 & 0.2613 & 0.4748 & 0.1146 \\

\midrule
\multicolumn{9}{c}{\textbf{GT$=$1}} \\
\cmidrule(lr){1-9}
A & 0.3918 & 0.8144 & 0.8144 & 0.5382
  & 0.0235 & 0.1197 & 0.3974 & 0.0587 \\
B & \textbf{0.5876} & \textbf{0.9278} & \textbf{0.9278} & \textbf{0.6883}
  & \textbf{0.0705} & \textbf{0.2457} & \textbf{0.5812} & \textbf{0.1334} \\
C & 0.3814 & 0.8041 & 0.8041 & 0.4907
  & 0.0192 & 0.1368 & 0.5192 & 0.0611 \\
D & 0.4124 & 0.8454 & 0.8454 & 0.5729
  & 0.0256 & 0.1517 & 0.4466 & 0.0699 \\
E & 0.1134 & 0.5258 & 0.5258 & 0.2431
  & 0.0192 & 0.1474 & 0.4829 & 0.0635 \\

\midrule
\multicolumn{9}{c}{\textbf{GT$>$1}} \\
\cmidrule(lr){1-9}
A & 0.0000 & 0.7222 & 0.3194 & 0.2374
  & 0.0570 & 0.4673 & 0.5051 & 0.1768 \\
B & \textbf{0.1111} & \textbf{0.9444} & \textbf{0.5880} & \textbf{0.3172}
  & \textbf{0.1156} & \textbf{0.6549} & \textbf{0.6270} & \textbf{0.2689} \\
C & 0.0000 & 0.7222 & 0.4282 & 0.2072
  & \textbf{0.1156} & 0.5226 & 0.5033 & 0.2405 \\
D & 0.1111 & 0.7222 & 0.3542 & 0.3003
  & 0.0955 & 0.5092 & 0.5147 & 0.2214 \\
E & 0.0556 & 0.5000 & 0.1667 & 0.1484
  & 0.1106 & 0.5260 & 0.5186 & 0.2395 \\

\bottomrule
\end{tabular}
}
\end{table}

Table~\ref{tab:ablation_unified} summarizes the ablation results under the type-constrained setting.
Variant B denotes the full QCEA model, while Variant A removes query conditioning, reducing the model to static matching based on source-side representations.

Overall, the full model achieves the most consistent performance across both datasets.
While some variants outperform on isolated metrics, such gains are not consistent, whereas the full model remains robust.

Removing query conditioning (A) leads to the largest and most consistent degradation, confirming that description-conditioned ranking is the primary source of improvement.
The performance drop is particularly evident on Herb, where ambiguity and candidate space are larger.

Removing the direction-aware Tucker projection (D) also causes a notable decline, especially on Herb (e.g., Recall@100 drops from 0.6069 to 0.4847), indicating the importance of direction-aware transformation for modeling non-bijective correspondence.

Eliminating graph propagation (C) results in consistent degradation, indicating that structural context provides complementary signals beyond textual semantics.

Removing residual fusion (E) further reduces performance in most settings.
The residual branch stabilizes representation learning and mitigates distortion from higher-order transformations.

These effects are particularly visible in challenging settings such as WM$\rightarrow$TCM and GT$>$1 cases.
For example, on Herb WM$\rightarrow$TCM, Recall@100 drops from 0.6302 to 0.4748 without residual fusion.

In summary, query conditioning provides the largest gain, followed by direction-aware Tucker transformation, while graph propagation and residual fusion contribute complementary improvements in robustness and stability.

\subsection{Training Dynamics and Performance Analysis}
\label{sec:training_analysis}
\begin{figure*}
\centering
\includegraphics[width=\textwidth,keepaspectratio]{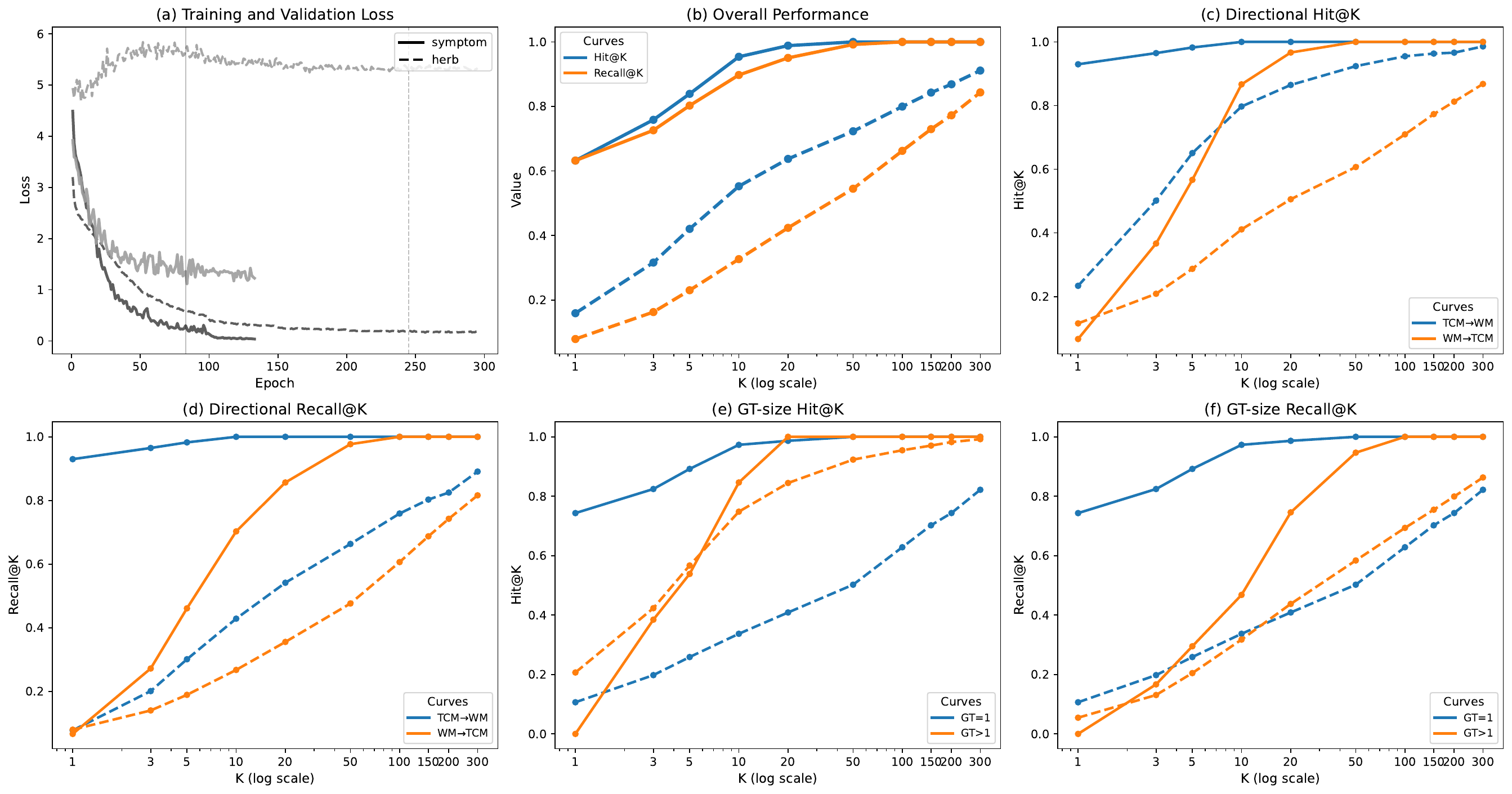}
\caption{
Training dynamics and retrieval performance on the Symptom and Herb datasets.
(a) Training and validation loss with selected best epochs.
(b) Overall Hit@K and Recall@K.
(c)-(d) Directional performance (TCM$\rightarrow$WM and WM$\rightarrow$TCM).
(e)-(f) Performance under different ground-truth cardinalities (GT$=$1 vs.\ GT$>$1).
Solid lines denote Symptom, dashed lines denote Herb, and $K$ is shown on a logarithmic scale.
}
\label{fig:training_curves}
\end{figure*}

Figure~\ref{fig:training_curves} shows stable optimization and consistent retrieval trends on both datasets.
The Symptom task converges earlier than the Herb task (best epoch 83 vs.\ 245), reflecting lower ambiguity and a smaller candidate space, whereas the Herb task requires more iterations due to its many-to-many and dispersed correspondence structure.

Figure~\ref{fig:training_curves}(b) further shows that retrieval performance generally improves with increasing $K$, but exhibits distinct saturation behaviors.
Symptom saturates early, indicating that correct targets are concentrated at top ranks, whereas Herb continues to benefit from larger $K$, reflecting more dispersed target distributions.

As shown in Figure~\ref{fig:training_curves}(c)--(d), directional asymmetry is consistently observed.
TCM$\rightarrow$WM outperforms WM$\rightarrow$TCM, especially at small $K$, suggesting that mapping from abstract to standardized representations is relatively easier.
The gap narrows as $K$ increases, indicating that correct matches remain present but are ranked lower in the more challenging direction.

Figure~\ref{fig:training_curves}(e)--(f) further shows that cardinality differentiates performance.
GT$=$1 cases saturate quickly, whereas GT$>$1 cases benefit more from larger $K$, particularly on Herb, highlighting the increased difficulty of multi-target alignment.

\subsection{Impact of Seed Alignment Ratio}
\begin{figure}
    \centering
    \includegraphics[width=\linewidth]{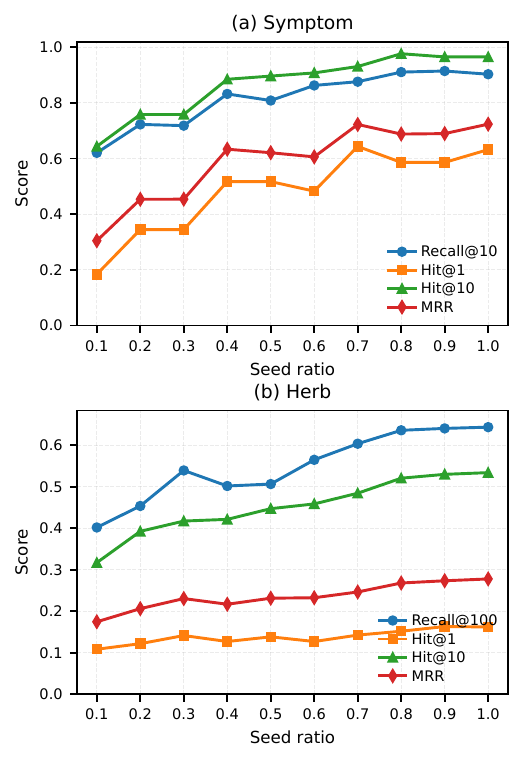}
    \caption{
    Impact of seed alignment ratio on (a) Symptom and (b) Herb tasks.
    Performance improves with increasing supervision.
    Recall@100 is used for Herb.
    }
    \label{fig:seed_ratio}
\end{figure}
Figure~\ref{fig:seed_ratio} shows the effect of varying the seed alignment ratio.
Performance generally improves with increasing supervision, with the largest gains in the low-resource regime (0.1--0.4) and diminishing returns thereafter, indicating that QCEA can establish meaningful alignment structure from limited seeds.

The impact of supervision differs across tasks.
Symptom achieves strong performance across all ratios, reflecting its relatively simple and near one-to-one correspondence.
In contrast, Herb benefits more from increased supervision, showing larger gains as the seed ratio increases, while remaining more challenging due to its larger candidate space and many-to-many structure.
Although Recall@100 is relatively high, lower Hit@1 and MRR indicate that precise ranking remains difficult.

Overall, QCEA scales effectively with increasing supervision and remains robust in low-resource settings, while the main remaining challenges arise from non-bijective and asymmetric correspondence.

\subsection{Impact on Downstream RAG}
\begin{figure*}
\centering
\includegraphics[width=\textwidth]{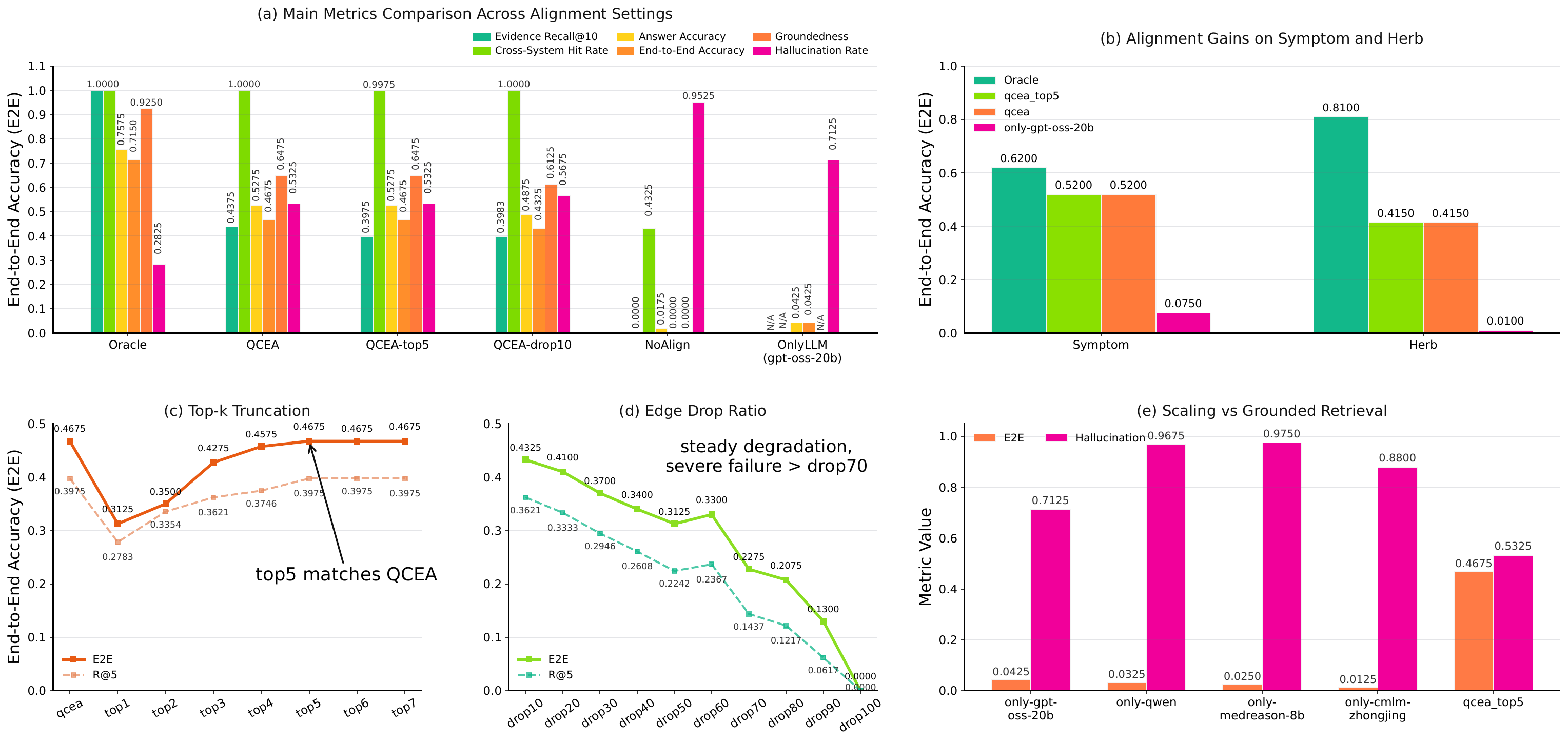}
\caption{Downstream RAG evaluation under different alignment settings.
(a) Overall comparison in terms of retrieval-level (evidence recall@K, cross-system hit rate), generation-level (answer accuracy), and end-to-end metrics (groundedness, end-to-end accuracy).
(b) Category-wise end-to-end accuracy for Symptom and Herb questions.
(c) Effect of confidence-based top-$k$ truncation of first-hop alignment candidates.
(d) Effect of random removal of first-hop alignment candidates.
(e) Comparison between retrieval-augmented settings and LLM-only baseline.}
\label{fig:rag}
\end{figure*}

Figure~\ref{fig:rag} examines the impact of alignment quality on downstream QA under the controlled RAG setting described in Section 4.
We analyze overall effectiveness, category-wise behavior, precision--coverage trade-offs, and the extent to which LLMs compensate for missing cross-system evidence.

\subsubsection{Overall comparison of downstream variants}

Fig.~\ref{fig:rag}(a) compares downstream QA performance under different alignment settings.
Oracle serves as an upper bound and achieves near-perfect retrieval and the strongest overall performance, confirming that correct first-hop alignment enables reliable and well-grounded reasoning.

Among practical variants, QCEA performs best overall, substantially improving retrieval, answer accuracy, and groundedness compared with all baselines.
Confidence-based truncation (the best-performing QCEA-TopX setting) slightly reduces retrieval quality but does not improve end-to-end performance, indicating limited benefit from aggressive candidate pruning.
In contrast, removing alignment (NoAlign) leads to near-zero performance, while LLM-only methods achieve very low accuracy, suggesting that parametric knowledge alone is insufficient to support cross-system reasoning.

These results suggest that accurate cross-system alignment is an important prerequisite for evidence-grounded QA, particularly when reasoning requires reliable cross-system evidence transfer.

\subsubsection{Category-wise analysis}

Fig.~\ref{fig:rag}(b) reports end-to-end accuracy for Symptom and Herb questions.
Under Oracle, Herb questions achieve higher accuracy, but this pattern reverses under predicted alignment.
With QCEA, Symptom achieves higher accuracy, while Herb performance drops more substantially relative to Oracle.

This suggests that Herb questions are more sensitive to alignment quality, likely due to their larger candidate space and many-to-many correspondence, whereas Symptom questions appear relatively more stable under imperfect alignment.

\subsubsection{Effect of truncation and candidate removal}

Fig.~\ref{fig:rag}(c)--(d) analyzes the impact of first-hop candidate control.
Aggressive top-$k$ truncation degrades performance, while moderate truncation yields a better balance between noise reduction and coverage.
In contrast, random removal of candidates leads to continuous degradation and near-collapse.

These results reveal a trade-off between precision and coverage: top-$k$ truncation can suppress noisy alignments, but excessive truncation breaks cross-system evidence chains required for downstream reasoning.
Strong degradation under random removal shows that candidate coverage is essential for preserving retrieval paths.
Together, these findings suggest that downstream performance depends not only on graph connectivity (coverage), but also on learned cross-system alignment quality, which determines whether relevant evidence can be surfaced during retrieval.

\subsubsection{LLM-only baselines and grounded evidence}
Fig.~\ref{fig:rag}(e) compares retrieval-based methods with LLM-only baselines.
All LLM-only variants perform substantially worse than retrieval-augmented methods, with lower answer accuracy and significantly higher hallucination rates.
Even with increased model capacity or domain specialization, LLM-only approaches remain unable to match the performance of alignment-enhanced RAG.

These results indicate that reliable performance depends on grounded cross-system evidence rather than parametric knowledge alone~\cite{mallen2022trust}.
Without explicit access to aligned cross-system evidence, LLMs fail to consistently retrieve and integrate relevant information for multi-hop reasoning, leading to unstable and ungrounded outputs.
Consequently, alignment-enhanced RAG improves reliability by grounding generation in aligned cross-system evidence.

\subsubsection{Case Study}

To further illustrate the mechanisms behind the quantitative results, we present a representative example.

\noindent\textbf{Query.} Which modern molecular components correspond to the TCM herb \emph{Mimenghua} (\emph{Buddlejae Flos})? (List up to 3.)

\noindent\textbf{NoAlign.} The retriever remains within the TCM subgraph and fails to access cross-system alignment edges, resulting in missing evidence and ungrounded answers.

\noindent\textbf{QCEA.} QCEA retrieves multiple cross-system candidates (e.g., \emph{Luteolin}, \emph{Apigenin}, \emph{Protocatechuic Acid}), enabling grounded reasoning and producing correct outputs, although the candidate set remains relatively broad.

\noindent\textbf{QCEA-TopX.} Truncation removes useful lower-ranked candidates, reducing retrieval coverage without improving answer quality, consistent with the precision--coverage trade-off observed in Fig.~\ref{fig:rag}(c).

\subsubsection{Summary of downstream RAG findings}
The downstream results support four main observations. 
First, accurate cross-system alignment appears essential for evidence-grounded QA.
Second, QCEA provides the strongest overall performance among practical settings, while truncation does not yield additional gains.
Third, the impact of alignment is category-dependent, with herb questions being more sensitive to alignment quality.
Fourth, preserving candidate coverage is critical, as both truncation and candidate removal degrade performance, whereas LLM-only methods remain far behind.

Overall, alignment quality consistently affects retrieval effectiveness, grounding, and end-to-end reliability in our experiments.

\section{Discussion and Implications}
The experimental results consistently indicate that cross-domain alignment quality plays a central role in downstream performance across multiple dimensions.
QCEA improves top-ranked metrics such as Hit@1, Hit@10, and MRR, particularly in the Symptom task, suggesting that query-conditioned alignment effectively concentrates relevant candidates near the top of the ranking.

The ablation study further highlights that query conditioning is the dominant factor driving performance, while direction-aware transformation and graph propagation provide complementary improvements, especially in more challenging settings with asymmetric mappings. Training dynamics reveal structural differences between tasks, with the Herb task exhibiting slower convergence due to its many-to-many correspondence nature.

Beyond alignment metrics, downstream RAG evaluation demonstrates that alignment quality consistently affects answer accuracy, grounding, and overall reliability. 
When alignment is degraded, the system exhibits reduced grounding and increased instability, whereas improved alignment leads to more consistent and evidence-based outputs.

These observations suggest that alignment can be viewed as a structural layer that influences how knowledge is accessed and organized for reasoning.
Rather than acting as a standalone reasoning mechanism, alignment shapes the candidate space from which reasoning systems draw evidence.
In this sense, improvements in alignment quality tend to propagate to downstream reasoning performance.

\paragraph*{Alignment as an upstream bottleneck.}
Our RAG experiments reveal that alignment acts as an upstream bottleneck: errors in alignment propagate directly into retrieval, and are difficult to correct by the generation model.
Under a controlled generation setting, the model (Qwen2.5-1.5B-Instruct) is unable to compensate for missing or misaligned cross-system evidence.
Consistently, improving alignment quality leads to better end-to-end performance, while degradation in alignment results in corresponding performance drops.

\paragraph*{Failure case analysis.}
Despite overall improvements, QCEA exhibits limitations in specific scenarios.
For herb--molecule alignment with GT$>$1, recall remains moderate (0.6270) compared to RDGCN (0.8811).
This occurs when a TCM herb corresponds to a large set of WM molecules with no clear semantic distinction in textual descriptions, leading to ranked but not highly concentrated predictions.
Additionally, WM$\rightarrow$TCM alignment remains challenging due to the inherent asymmetry in abstraction levels between the two medical systems.

\textbf{Limitations.} 
Our conclusions are based on empirical evaluation and do not by themselves establish causal relationships between alignment and downstream reasoning.
Additionally, experiments are conducted in a transductive setting and focus on TCM--WM integration, which may limit generalization to other domains.
Finally, the use of a lightweight language model in RAG evaluation emphasizes the impact of alignment quality, but results may vary with larger models.

\textbf{Implications.}
These findings suggest that improving cross-domain alignment is a practical and effective direction for enhancing knowledge integration systems.
Future work may explore hybrid approaches that combine structured alignment with generative models, as well as extensions to broader multi-domain knowledge settings.

\section{Conclusions}
We presented QCEA, a query-conditioned entity alignment framework for bridging heterogeneous TCM and WM knowledge graphs.
By reformulating alignment as a query-conditioned ranking problem, QCEA captures context-dependent, asymmetric, and non-bijective correspondence in cross-system medical knowledge.
Experiments on SymMap-derived datasets demonstrate consistent improvements over representative baselines, with notable gains in semantically ambiguous settings and rank-sensitive retrieval scenarios.
Further downstream RAG evaluation shows that improved alignment enhances the availability and ranking of cross-system evidence, leading to improved retrieval effectiveness, grounding, and end-to-end reliability, while reducing unsupported or inconsistent outputs in cross-system medical QA.

Overall, these results suggest that query-conditioned alignment provides an effective and practical approach for integrating heterogeneous medical knowledge and supporting knowledge-grounded reasoning. 
Future work will explore incorporating richer clinical context, improving robustness under sparse supervision, and extending alignment to more complex multi-hop reasoning settings.

\section*{Data Availability}
The data used in this study are derived from the publicly available SymMap database.
The processed benchmark datasets and evaluation splits are available from the corresponding author upon reasonable request.

\section*{Declaration of Competing Interest}
The authors declare that they have no known competing financial interests or personal relationships that could have appeared to influence the work reported in this paper.

\bibliographystyle{cas-model2-names}
\bibliography{references}
\end{document}